%% file: emnlp-ijcnlp-2019.tex
\documentclass[11pt,a4paper]{article}
\usepackage[hyperref]{emnlp-ijcnlp-2019}

\usepackage{times}
\usepackage{latexsym}
\usepackage{subcaption}

\usepackage{url}
\usepackage{algorithm}
\usepackage[noend]{algpseudocode}

\input{preamble}

\aclfinalcopy

\title{Metric Learning for Dynamic Text Classification}

\author{Jeremy Wohlwend\ \ \ \ \ Ethan R. Elenberg\ \ \ \ \ Samuel Altschul\ \ \ \ \ Shawn Henry\ \ \ \ \ Tao Lei \vspace{0.1in} \\ 
ASAPP, Inc. \vspace{0.1in} \\ 
{\tt \{jeremy,eelenberg,saltschul,shawn,tao\}@asapp.com}}

\date{}

\begin{document}
\maketitle
\begin{abstract}
  \input{abstract}
\end{abstract}

\section{Introduction}\label{sec:intro}

\input{intro}

\section{Related Work}
\input{related_work.tex}

\section{Model Framework}\label{sec:framework}
\input{framework}

\section{Experiments} \label{sec:experiments}
\input{experiments}

\section{Analysis} \label{sec:analysis}
\input{analysis}

\section{Conclusions}
\input{conclusions}

\bibliography{emnlp-ijcnlp-2019}

\bibliographystyle{emnlp_natbib}

\clearpage

\appendix

\section{Proof of Proposition~\texorpdfstring{\ref{prop:hyp_mean}}{\ref*{prop:hyp_mean}}}\label{sec:hyp_mean_proof}
\input{mean_proof}

\section{Two Differing Means}\label{sec:means_differ_example}
\input{means_differ_example}

\section{20 Newsgroups Hierarchy}\label{sec:app_newsgroups_hier}

\input{hierarchy_appendix}

\newpage \newpage \section{Dataset Details}\label{sec:app_datasets}
\input{appendix_datasets}

\section{Additional Experiments}\label{sec:app_experiments}
\input{extra_experiments}

\end{document}

%% file: preamble.tex
\usepackage{nicefrac}
\usepackage{tikz}
\usetikzlibrary{positioning}
\usetikzlibrary{arrows.meta}
\usepackage{amsmath}
\usepackage{amsthm}
\usepackage{amssymb}

\newtheorem{proposition}{Proposition}
\newtheorem*{proposition_nonum}{Proposition}

\DeclareMathOperator*{\argmin}{arg\,min}
\DeclareMathOperator*{\arccosh}{arccosh}

\newcommand{\hypspace}[0]{\mathbb{H}}

%% file: abstract.tex
Traditional text classifiers are limited to predicting over a fixed set of labels. However, in many real-world applications the label set is frequently changing. For example, in intent classification, new intents may be added over time while others are removed.

We propose to address the problem of dynamic text classification by replacing the traditional, fixed-size output layer with a learned, semantically meaningful metric space. Here the distances between textual inputs are optimized to perform nearest-neighbor classification across overlapping label sets. Changing the label set does not involve removing parameters, but rather simply adding or removing support points in the metric space. Then the learned metric can be fine-tuned with only a few additional training examples.

We demonstrate that this simple strategy is robust to changes in the label space. Furthermore, our results show that learning a non-Euclidean metric can improve performance in the low data regime, suggesting that further work on metric spaces may benefit low-resource research. \footnote{Code: github.com/asappresearch/dynamic-classification}

%% file: intro.tex
\begin{figure}[!t!]
    \centering
    \includegraphics[width=2.8in]{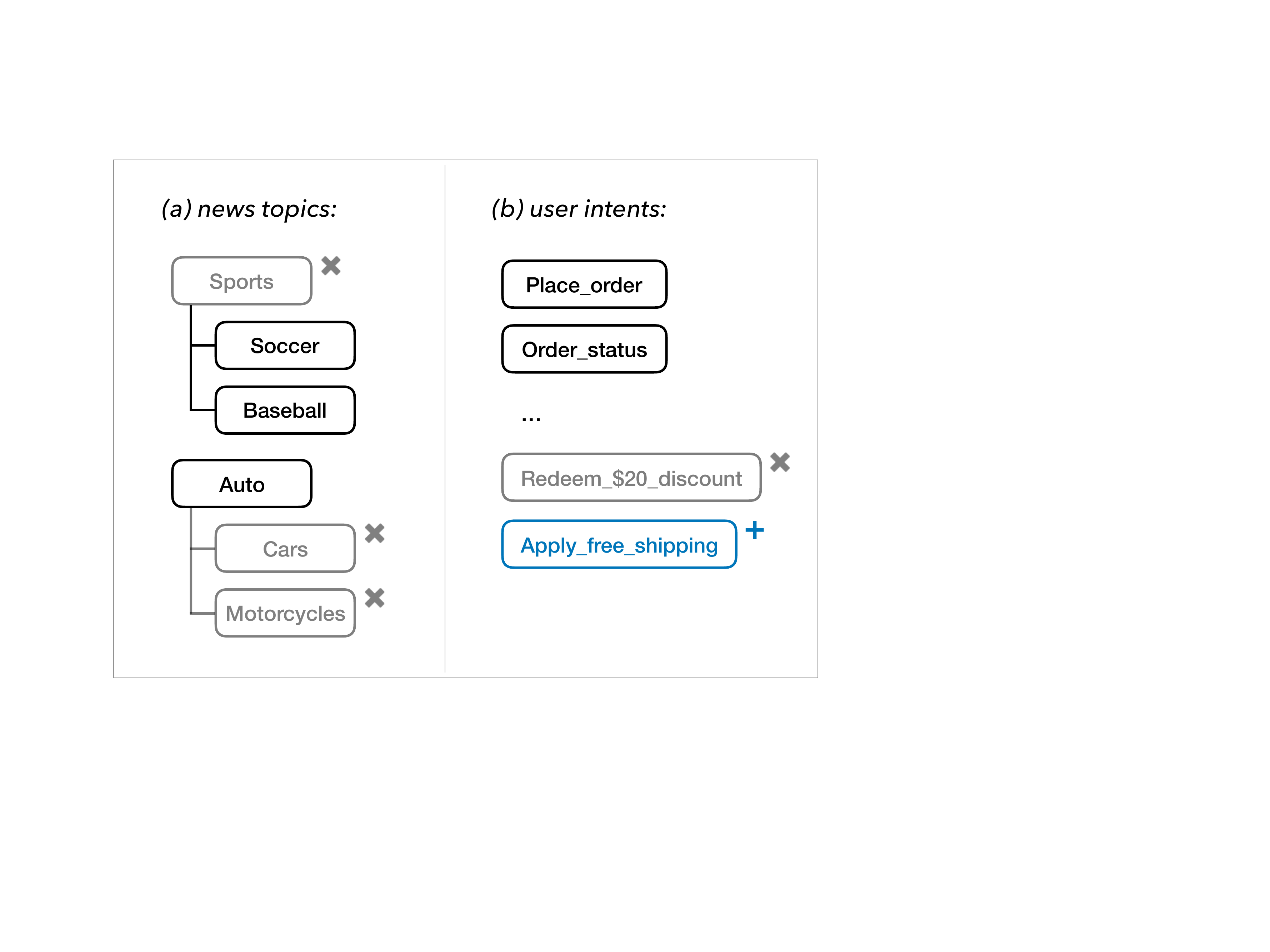}
    \caption{\label{fig:example_hierarchy}Examples of dynamic classification. In the hierarchical setting (left), new labels are created by splitting and merging old labels. In the flat setting (right), arbitrary labels can be added or removed.}
\end{figure}

Text classification often assumes a static set of labels.
While this assumption holds for tasks such as sentiment analysis and part-of-speech tagging \cite{panglee-sentiment, kim-classification, brants2000tnt, collins2002discriminative, toutanova2003feature}, it is rarely true for real-world applications.
Consider the example of news categorization in Figure~\ref{fig:example_hierarchy}~(a).
A domain expert may decide that the \textit{Sports} class should be separated into two distinct \textit{Soccer} and \textit{Baseball} sub-classes, and conversely merge the two \textit{Cars} and \textit{Motorcycles} classes into a single \textit{Auto} category.
Another example is user intent classification in task-oriented dialog systems.
In Figure~\ref{fig:example_hierarchy}~(b) for example, an intent to redeem a reward can be removed when the option is no longer available, while a new intent to apply free shipping can be added to the system.

In all of these applications, the classifier must remain applicable for \emph{dynamic classification}, a task where the label set is rapidly evolving.

Several factors make the dynamic classification problem difficult.
First, traditional classifiers are not suited to changes in the label space. These classifiers produce a fixed sized output which aligns each of the dimensions to an existing label. Thus, adding or removing any label requires changing the model architecture.
Second, while it is possible to retain some model parameters, such as in hierarchical classification models, these architectures must still learn separate weights for every new class or sub-class \cite{cai2004hierarchical, kowsari2017hdltex}.
This is problematic because the new class labels often come with very few training examples, providing insufficient information for learning accurate model weights. Furthermore, these models do not leverage information across similar labels, which weakens their ability to adapt to new target labels~\cite{kowsari2017hdltex, tsochantaridis2005large, cai2004hierarchical}.

 We propose to address these issues by learning an embedding function which maps input text into a semantically meaningful metric space. The parameterized metric space, once trained on an initial set of labeled data, can be used to perform classification in a nearest-neighbor fashion (by comparing the distance from the input text to reference texts with known label). As a result, the classifier becomes agnostic to changes in the label set. One remaining design challenge, however, is to learn a representation that best leverages the relationship between old and new labels. In particular, the label split example in Figure~\ref{fig:example_hierarchy}~(b) shows that new labels are often formed by partitioning an old label. This suggests that the classifier may benefit from a metric space that can better represent the structural relationships between labels. Given the hierarchical relationship between the old and new labels, we choose a space of negative curvature (hyperbolic), which has been shown to better embed tree-like structure ~\cite{nickel2017poincare,sala2018representation,gu2019mixed}. 
 
 Our two main contributions are outlined below:
 \begin{enumerate}
 \item We design an experimental framework for \textit{dynamic text classification}, and propose a classification strategy based on prototypical networks, a simple but powerful metric learning technique~\cite{snell2017prototypical}.

\item We construct a novel prototypical network adapted to hyperbolic geometry. This requires deriving useful prototypes to represent a set of points on the negatively curved Riemannian manifold. We state sufficient theoretical conditions for the resulting optimization problem to converge. To the best of our knowledge, this is the first application of hyperbolic geometry to text classification beyond the word level.

\end{enumerate}

We perform a thorough experimental analysis by considering the model improvements across several aspects -- low-resource fine-tuning, impact of pretraining, and ability to learn new classes. We find that the metric learning approach adapts more gracefully to changes in the label distribution, and outperforms traditional, fixed size classifiers in every aspect of the analysis. Furthermore, our proposed hyperbolic prototypical network outperforms its Euclidean counterpart in the low-resource setting, when fewer than 10 examples per class are available.

%% file: related_work.tex
\paragraph{Prototypical Networks and Manifold Learning:}
This paper builds on the prototypical network architecture~\cite{snell2017prototypical}, which was originally proposed in the context of few-shot learning. In both their work and ours, the goal is to embed training data in a space such that the distance to \textit{prototype} centroids of points with the same label define good decision boundaries for labeling test data with a nearest neighbor classifier.
Building on earlier work in metric learning~\cite{vinyals2016matchingnet, Sachin2017}, the authors show that learned prototype points also help the network classify inputs into test classes for which minimal data exists.

This architecture has found success in computer vision applications such as image and video classification~\cite{kilian, ustinova2016histogram, luo_nips17_label}.
Very recently, prototypical network architectures have shown promising results on relational classification tasks~\cite{han2018fewrel,gao2019rc}.
To the best of our knowledge, our work is the first application of prototypical network architectures to text classification using non-Euclidean geometry.\footnote{\citet{snell2017prototypical} discuss their formulation in the context of Euclidean distance, cosine distance (spherical manifold), and general Bregman divergences; however, classical Bregman divergence does not easily generalize to hyperbolic space (Section~\ref{subsec:distance}).}

Concurrent with the writing of this paper, \cite{khrulkov2019hyp_image} applied several hyperbolic neural networks to few-shot image classification tasks. However, their prototypical network uses the Einstein midpoint rather than the definition of the mean we use in Section~\ref{subsec:distance}. 

In \cite{chen2019interaction} the authors embed the labels and data separately, then predict hierarchical class membership using an interaction model. Our model directly links embedding distances to model predictions, and thus learns an embedded space that is more amenable to low-resource, dynamic classification tasks.

Hyperbolic geometry has been explored in classical works of differential geometry~\cite{thurston-hyperbolic, cannon-hyperbolic, berger-riemanian}. More recently, hyperbolic space has been studied in the context of developing neural networks with hyperbolic parameters~\cite{eth-hnn}. 

In particular, recent work has successfully applied hyperbolic geometry to graph embeddings~\cite{sarkar, nickel2017poincare, nickel2018lorentz, sala2018representation, eth-entailment,gu2019mixed}. In all of these prior works, the model's parameters correspond to node vectors in hyperbolic space that require Riemannian optimization. In our case, only the model's outputs live in hyperbolic space---not its parameters, which avoids propagating gradients in hyperbolic space and facilitates optimization. This is explained in more detail in Section~\ref{subsec:distance}.

\paragraph{Hierarchical or Few-shot Text Classification:}
Many classical models for multi-class classification incorporate a hierarchical label structure~\cite{tsochantaridis2005large, cai2004hierarchical, yen2016pdsparse, naik2013multitask,sinha2018}.
Most models proceed in a top-down manner: a separate classifier (logistic regression, SVM, etc.) is trained to predict the correct child label at each node in the label hierarchy.

For instance, HDLTex~\cite{kowsari2017hdltex} addresses large hierarchical label sets explicitly by training a stacked, hierarchical neural network architecture.
Such approaches do not scale well to deep and large label hierarchies, while our method can adapt to more flexible settings, such as adding or removing labels, without adding extra parameters.

Our work also relates to text classification in a low-resource setting. While a wide range of methods improve accuracy by leveraging external data such as multi-task training~\cite{miyato2016adversarial,adan-chen-tacl,yu2018,guo2018}, semi-supervised pretraining~\cite{dai2015semi}, and unsupervised pretraining~\cite{Peters:2018,devlin2018bert}, our method makes use of the structure of the data via metric learning.
As a result, our method can be easily combined with any of these methods to further improve model performance.

%% file: framework.tex
This section provides the details of each component of our framework, starting with a more detailed formulation of \textit{dynamic classification}. We then provide some background on prototypical networks, before introducing our hyperbolic variant and its theoretical guarantees.

\subsection{Dynamic Classification}

Formally, we formulate \textit{dynamic classification} as the following problem: given access to an old, labeled training corpus $(x_{i}, y_{i}) \in \mathcal{X}_{old} \times \mathcal{Y}_{old}$, we are interested in training a classifier $h\!: \mathcal{X}_{new} \mapsto \mathcal{Y}_{new}$ with a few examples $(x_{j}, y_{j}) \in \mathcal{X}_{new} \times \mathcal{Y}_{new}$. Unlike few-shot learning, the old and new datasets need not be disjoint ($\mathcal{X}_{old} \cap \mathcal{X}_{new} \neq \emptyset$, $\mathcal{Y}_{old} \cap \mathcal{Y}_{new} \neq \emptyset$).

We consider two different cases: 1) new labels arrive as a consequence of new input data $\mathcal{X}_{new} \setminus \mathcal{X}_{old}$, and 2) during label splitting/merging, some new examples may be constructed by relabeling old examples from $y_i \in \mathcal{Y}_{old}$ to $y_j \in \mathcal{Y}_{new} \setminus \mathcal{Y}_{old}$. This latter case is of particular interest as the classifier may be able to leverage its knowledge of old labels in learning to classify new ones. 

There are many natural approaches to this problem. First, a fixed model trained on $\mathcal{X}_{old} \times \mathcal{Y}_{old}$ may be applied directly to classify examples in $\mathcal{X}_{new} \times \mathcal{Y}_{new}$, which we refer to as an \textit{un-tuned} model. Alternately, a pretrained model may also be fine-tuned on $\mathcal{X}_{new} \times \mathcal{Y}_{new}$. Finally, it is also possible to train from scratch on $\mathcal{X}_{new} \times \mathcal{Y}_{new}$, disregarding the model weights trained on the old data distribution. We compare strategies in Sections~\ref{sec:experiments}--\ref{sec:analysis}.

\subsection{Episodic Training}
The standard prototypical network is trained using episodic training, as described in \cite{snell2017prototypical}. We view our model as an embedding function which takes textual inputs and outputs points in the metric space.

Let $d(x,y)$ denote the distance between two points $x$ and $y$ in our metric space, and let $f$ denote our embedding function. At each iteration, we form a new episode by sampling a set of target labels, as well as support and query points for each of the sampled labels. Let $N_C$, $N_S$, and $N_Q$, be the number of classes tested, the number of support points used, and the number of query points used in each episode, respectively. 

For each episode, we first sample $N_C$ classes, $C = \{c_i| i=1,\ldots,N_C \}$, uniformly across all training labels. We then build a set of support points $S_i= \{s_{i,j} | j =1,\ldots,N_S \}$ for each of the selected classes by sampling $N_S$ training examples from each selected class. For each support set, we compute a prototype vector $p^*_i$. For the standard Euclidean prototypical network, we use the mean of the embedded support set:
\begin{align}
    p^*_i = \frac{1}{N_S}\sum_{j=1} ^{N_S} f(s_{i,j}) \enspace . \label{eq:euclidean_mean}
\end{align}

To compute the loss for an episode, we further sample $N_Q$ query points $Q= \{x_{i,j} | j=1, \ldots, N_Q\}$ which do not appear in the support set of the episode, for each selected class $c_i$. We then encode each query sequence and apply a softmax function over the negative distances from the query points to the episode's class prototypes. This yields a probability distribution over classes, and we take the negative log probability of the true class, averaged over the query points, to get the loss for the episode.
\begin{align*}
\begin{split}
    \frac{-1}{N_C N_Q}\sum_{i=1}^{N_C} \sum_{j=1}^{N_Q}  \log\!\left[\frac{\exp(-d(f(x_{i,j}), p^*_i))}{\sum_{k}\exp(-d(f(x_{i,j}), p^*_{k}))}\right] ,
\end{split}
\end{align*}

where $k$ in the denominator ranges from $1$ to $N_C$. The steps of a single episode are summarized in Algorithm \ref{alg:prototypical_alg}.

Once episodic training is finished, the prototype vectors for a class can be computed as the mean of the embeddings of any number of items in the class. In our experiments, we use the whole training set to compute the final class prototypes, but under lower resources, fewer support points could also be used.

\begin{algorithm}[!t!]

\caption{Prototypical Training Episode}\label{alg:prototypical_alg}
\begin{flushleft} 
\textbf{Input:} $D$ -- set of $(x, y)$ pairs \\
$D_i$ -- all pairs with $y=i$ \\
$N_C$ -- number of classes sampled each episode\\
$N_S$ -- number of support points \\
$N_Q$ -- number of query points
\end{flushleft}
\begin{algorithmic}[1]
\Procedure{Episode}{$D$, $N_C$,$N_S$, $N_Q$}
\State $C\gets \textsc{Sample}(D, N_C)$

\For{$i \in C$}
\State $S_i$ $\gets$ $\textsc{Sample}(D_i, N_S)$
\State $Q_i$ $\gets$ $\textsc{Sample}(D_i \setminus S_i, N_Q)$
\State $c_i$ $\gets$ $\textsc{Prototype}(S_i)$
\EndFor
\State $P$ $\gets$ \textsc{Concat}($c_0$; $c_1$; ...; $c_{N_C}$)
\State $\operatorname{Loss}$ $\gets$ 0
\For{ each $Q_i$}
\State $d_i\gets \textsc{PairwiseDist}(Q_i, P)$
\State $\operatorname{Loss}\gets \operatorname{Loss} - \frac{1}{N_C N_Q}\log\left[\frac{e^{-d_i}}{\sum_j e^{-d_j}}\right]$
\EndFor 

\EndProcedure
\end{algorithmic}
\end{algorithm}

\subsection{Hyperbolic Prototypical Networks}\label{subsec:distance}
In this section we discuss the hyperbolic prototypical network which can better model structural relationships between labels. We first review the hyperboloid model of hyperbolic space and its distance formula. Then we describe the main technical challenge of computing good prototypes in hyperbolic space. Proofs of our uniqueness and convergence can be found in the Appendix ~\ref{sec:hyp_mean_proof}.
We also describe a second, \textit{distinct} method for computing prototypes which is used to initialize our main method during experiments. A detailed discussion of this point is provided in Appendix~\ref{sec:means_differ_example}.

Hyperbolic space can be interpreted as a continuous analogue of a tree~\cite{cannon-hyperbolic, krioukov2010networks}. While trees on $n$ vertices can be embedded in Euclidean space with $\log(n)$ dimensions with minimal distortion, hyperbolic space needs only 2 dimensions. Additionally, the circumference of a hyperbolic disk grows exponentially with its radius. Therefore, hyperbolic models have room to place many prototypes equidistant from a common parent while maintaining separability from other classes. We argue that this property helps text classification with latent hierarchical structures (e.g. dynamic label splitting).

The reader is referred to Section 2.6 of~\cite{thurston-hyperbolic} for a detailed introduction to hyperbolic geometry, and to \cite{cannon-hyperbolic} for a more gentle introduction. In this section we have adopted the sign convention of~\cite{sala2018representation}.

Hyperbolic space in $d$ dimensions is the unique, simply connected, $d$-dimensional, Riemannian manifold with constant curvature $-1$. The hyperboloid (or Lorentz) model realizes $d$-dimensional hyperbolic space as an isometric embedding inside $\mathbb{R}^{d+1}$ endowed with a signature $(1, d)$ bilinear form.

Specifically, let the coordinates of any $a \in \mathbb{R}^{d+1}$ be $a = (a_0, a_1, ..., a_d)$.  Then we can define a bilinear form on $\mathbb{R}^{d+1}$ by 
\begin{align}
B(x,y) = x_0 y_0 - \sum_{j=1}^d x_j y_j \enspace, \label{eq:hyperboloid_bilinear}
\end{align}

which allows us to define the hyperboloid to be the set $\{x \in \mathbb{R}^{d+1}| B(x, x) = 1 \text{ and } x_0 > 0\}$. We induce a Riemannian metric on the hyperboloid by restricting $B(\cdot, \cdot)$ to the hyperboloid's tangent space. The resulting Riemannian manifold is hyperbolic space $\mathbb{H}^d$. For $x, y \in \hypspace^d$ the hyperbolic distance is given by
\begin{align}
d_{\hypspace}(x, y) = \arccosh( B(x, y)) \label{eq:hyperboloid_distance}.
\end{align}

There are several equivalent ways of defining hyperbolic space. We choose to work primarily in the hyperboloid model over other models (\textit{e.g.} Poincar{\'e} disk model) for improved numerical stability. We use the $d$-dimensional output vector $h$ of our network and project it on the hyperboloid embedded in $d+1$ dimensions:
\begin{align}
h_0 = \sqrt{\sum_{i=1}^d h_i^2 + 1} \enspace, \qquad
\bar{h} = [h_0; h] \enspace .
\label{eq:projection}
\end{align}

A key algorithmic difference between the Euclidean and the hyperbolic model is the computation of prototype vectors.
There are multiple definitions that generalize the notion of a mean to general Riemannian manifolds.

One sensible mean $p^{\star}_{X}$ of a set $X$ is given by the point which minimizes the sum of squared distances to each point in $X$.

\begin{align}
\begin{split}
p^{\star}_{X} &= \argmin_{p \in \hypspace^d} \phi_{X}(p) \\
&= \argmin_{p \in \hypspace^d} \sum_{x \in X} d_{\hypspace^d}(p, x)^2 \enspace . \label{eq:mean_function}
\end{split}
\end{align}

A proof for the following proposition can be found in Appendix ~\ref{sec:hyp_mean_proof}. We note that concurrent with the writing of this paper, a generalized version of our result appeared in~\cite{gu2019mixed} as Lemma 2.

\begin{proposition} \label{prop:hyp_mean}
Every finite collection of points $X$ in $\hypspace^d$ has a unique mean $p^{\star}_{X}$. Furthermore, solving the optimization problem \eqref{eq:mean_function} with Riemannian gradient descent will converge to $p^{\star}_{X}$.
\end{proposition}

In an effort to derive a closed form for $p^{\star}_{X}$ (rather than solve a Riemannian optimization problem), we found the following expression to be a good approximation. It is computed by averaging the vectors in $X$ and scaling them by the constant which projects this average back to the hyperboloid:
\begin{align}
\begin{split}
\hat{p} = \frac{1}{|X|}\sum_{x \in X} x , \quad
\tilde{p} = \frac{\hat{p}}{\sqrt{B(\hat{p},\hat{p})}} .
    \label{eq:closed_form_mean}
    \end{split}
\end{align}
$p^{\star}_{X} \neq \tilde{p}$ can be shown to differ through a simple counterexample, although in practice we find little difference between their values during experiments. The proof is be provided in Appendix \ref{sec:means_differ_example}.

\subsection{Implementation and Stability}
Our final hyperbolic prototypical model combines both definitions with the following heuristic: initialize problem~\eqref{eq:mean_function} with $\tilde{p}$ and then run several iterations of Riemannian gradient descent. We find that it is possible to backpropagate through a few steps of the gradient descent procedure described above during prototypical model training. However, we also find that the model can be trained successfully when detaching the gradients with respect to the support points. This suggests that prototypical models can be trained in metric spaces where the mean or its gradient cannot be computed efficiently. Further experimental details are provided in the next section.

Our prototypical network loss function uses both squared Euclidean distance and squared hyperbolic distance for similar reasons. Namely, the distance between two close points is much less numerically stable than the squared distance. In the Euclidean case, the derivative of $\sqrt{s}$ is undefined at zero. In the hyperbolic case, the derivative of $\arccosh(s)$ at $1$ is undefined, and $B(x,x) = 1$ for points on the hyperboloid. If we instead use the \textit{squared} hyperbolic distance, L'H\^{o}pital's rule implies that the derivative of $\arccosh(b)^2$ as $b \rightarrow 1+$ is $2$, allowing gradients to backpropagate through the squared hyperbolic distance without issue.

%% file: experiments.tex
We evaluate the performance of our framework on several text classification benchmarks, two of which exhibit a hierarchical label set. We only use the label hierarchy to simulate the label splitting discussed in Figure 1 (a). The models are not trained with explicit knowledge of the hierarchy, as we assume that the full hierarchy is not known \textit{a priori} in the dynamic classification setting. A description of the datasets is provided below:

\begin{itemize}
\item 20 Newsgroups (NEWS): This dataset is composed of nearly 20,000 documents, distributed across 20 news categories. We use the provided label hierarchy to form the depth $3$ tree used throughout our experiments. We use 9,044 documents for training, 2,668 for validation, and 7,531 for testing.
\item Web of Science (WOS): This dataset was used in two previous works on hierarchical text classification~\cite{kowsari2017hdltex, sinha2018}. It contains 134 topics, split across 7 parent categories. It contains 46,985 documents collected from the Web of Science citation index. We use 25,182 documents for training, 6,295 for validation, and 15,503 for testing.  

\item Twitter Airline Sentiment (SENT): This dataset consists of public tweets from customers to American-based airlines labeled with one of $10$ reasons for negative sentiment (\textit{e.g.} Late Flight, Lost Luggage).\footnote{\url{https://www.kaggle.com/crowdflower/twitter-airline-sentiment}} We preprocess the data by keeping only the negative tweets with confidence over $60\%$. This dataset is non-hierarchical and composed of nearly 7500 documents. We use 5,975 documents for training, 742 for validation, and 754 for testing.
\end{itemize}

\begin{table*}[t!]
	\begin{center}
		\begin{tabular}{|l|l|r|r|r|r|}
		    \hline
			Dataset & Model & $n_{fine}=5$ &  $n_{fine}=10$ &   $n_{fine}=20$ &  $n_{fine}=100$ \\
			\hline
			    & MLP   &  $37.3 \pm 2.9$ & $43.8 \pm 3.5$ &  $45.7 \pm 3.8$ &   $57.4 \pm 3.5$ \\ 
			SENT & EUC  & $39.6 \pm 6.4$ & $45.5 \pm 1.8$ & $47.7 \pm 4.7$ &   $\bf{62.7 \pm 2.1}$ \\
			    & HYP   & $\bf{42.2 \pm 3.5}$ &  $\bf{47.1 \pm 4.8}$  &  $\bf{53.0 \pm 2.3}$ &   $\bf{62.7 \pm 2.2}$ \\ \hline
			\hline
			    & MLP & $49.2 \pm 1.0$ &  $55.9 \pm 2.5$ &  $68.5  \pm 1.1$ &  $76.3 \pm 0.5$ \\
			NEWS & EUC & $56.5 \pm 0.4$ & $65.6 \pm 1.0$ & $\bf{74.2 \pm 0.6}$ &   $\bf{79.8 \pm 0.2}$ \\
			    & HYP & $\bf{64.8 \pm 2.8}$& $\bf{69.7 \pm 1.0}$ & $72.9 \pm 0.5$ & $78.8 \pm 0.4$ \\ \hline
			    & MLP &   $36.6 \pm 1.1$ & $46.8 \pm 1.2$ & $62.8 \pm 0.6$ &  $68.9 \pm 0.5$\\
			WOS & EUC &    $49.4 \pm 1.0$ & $59.2 \pm 0.4$ & $\bf{70.4 \pm 0.4}$ & $73.3 \pm 0.2$ \\
			    & HYP & $\bf{54.5 \pm 1.4}$ & $\bf{60.7 \pm 0.9}$ &  $70.2 \pm 0.7$ & $\bf{73.5 \pm 0.5}$ \\
			 \hline
	    \end{tabular}
	\end{center}
    \caption{Test accuracy for each dataset and method. Columns indicate the number of examples per label $n_{fine}$ used in the fine tuning stage. In all cases, the prototypical models outperform the baseline. The hyperbolic model performs best in the low data regime, but both metrics perform comparably when data is abundant.}\label{tab:dynamic_full}
\end{table*}

\paragraph{Dynamic Setup:} We construct training data for the task of dynamic classification as follows. First, we split our training data in half. The first half is used for pretraining and the second for fine-tuning. To simulate a change in the label space, we randomly remove $p>0$ fraction of labels in the pretraining data. This procedure yields two label sets, with $\mathcal{Y}_{old}$ (pretraining)  $\subset \mathcal{Y}_{new}$ (fine-tuning). In our experiments, we further vary the amount of data available in the fine-tuning set.
For the flat dataset, the labels to be removed are sampled uniformly. In the hierarchical case, we create $\mathcal{Y}_{old}$ by randomly collapsing leaf labels into their parent classes, as shown previously in Figure \ref{fig:example_hierarchy}.  

\paragraph{Hyperparameters and Implementation Details:} We apply the same encoder architecture throughout all experiments. We use a 4 layer recurrent neural network, with SRU cells~\cite{lei2018sru} and a hidden size of 128. We use pretrained GloVe embeddings~\cite{pennington2014glove}, which are fixed during training. A sequence level embedding is computed by passing a sequence of word embeddings through the recurrent encoder, and taking the embedding for the last token to represent the sequence. We use the ADAM optimizer with default learning of 0.001, and train for 100 epochs for the baseline models and 10,000 episodes for the prototypical models, with early stopping. In our experiments, we use $N_S = 4$, $N_Q = 64$. We use the full label set every episode for all datasets except WOS, for which we use $N_C=50$. We use a dropout rate of 0.5 on NEWS and SENT, and 0.3 for the larger WOS dataset. We tuned the learning rate and dropout for each model on a held-out validation set.

For the hyperbolic prototypical network, we follow the initialization and update procedure outlined at the end of Section~\ref{subsec:distance} with $5$ iterations of  Riemannian gradient descent during training and $100$ iterations during evaluation.

We utilize negative squared distance in the softmax computation in order to improve numerical stability. The means are computed via~\eqref{eq:mean_function} during both training and model inference. However, this computation is treated as a constant during backpropagation as described in
Section~\ref{subsec:distance}.

\paragraph{Baseline:} Our baseline model consists of the same recurrent encoder and an extra linear output layer which computes the final probabilities over the target classes. 

In order to fine-tune this multilayer perceptron (MLP) model on a new label ontology, we reuse the encoder, and learn a new output layer. This differs from the prototypical models for which the architecture is kept unchanged.

\paragraph{Evaluation:} We evaluate the performance of our models using accuracy with respect to the new label set $\mathcal{Y}_{new}$. We also highlight accuracy on only the classes introduced during label addition/splitting, \textit{i.e.} $\mathcal{Y}_{new} \setminus \mathcal{Y}_{old}$. All results are averaged over $5$ random label splits with $p=0.3$.

\paragraph{Results:} Table~\ref{tab:dynamic_full} shows the accuracy of the fine tuned models for all three methods. The SENT dataset shows performance in the case where completely new labels are added during fine tuning. In the NEWS and WOS datasets new labels originate from the splits of old labels. 

In all cases, the prototypical models outperform the baseline MLP model significantly, especially when the data in the new label distribution is in the low-resource regime (+5--15\% accuracy).
We also see an increase in performance in the high data regime of up to 5\%.

Table~\ref{tab:dynamic_full} further shows that the hyperbolic model outperforms its Euclidean counterpart in the low data regime on the NEWS and WOS datasets. This is consistent with our hypothesis (and previous work) that hyperbolic geometry is well suited for  hierarchical data. Interestingly, the hyperbolic model also performs better on the non-hierarchical SENT dataset when given few examples, which implies that certain metric spaces may be generally stronger in the low-resource setting. In the high data regime, however, both prototypical models perform comparably.

%% file: analysis.tex
In this section, we examine several aspects of our experimental setup more closely, and use the SENT and NEWS datasets for this analysis. Results on WOS can be found in the Appendix ~\ref{sec:app_experiments}.

\paragraph{Benefits of Pretraining}
\begin{figure}[!ht]
    \centering
    \begin{subfigure}{\columnwidth}
    \centering
    \includegraphics[width=\textwidth]{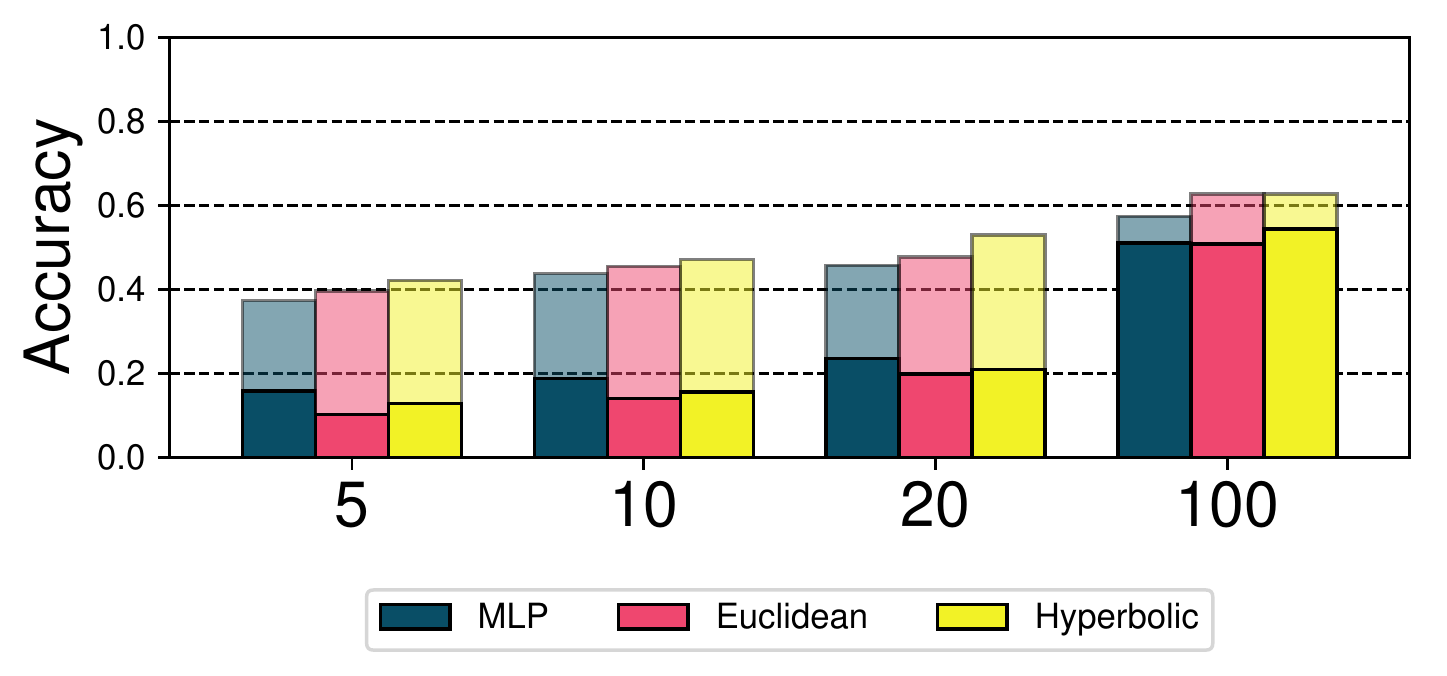}
    \caption{SENT \label{fig:acc_trec}}
    \end{subfigure}\\
    \begin{subfigure}{\columnwidth}
    \centering
    \includegraphics[width=\textwidth]{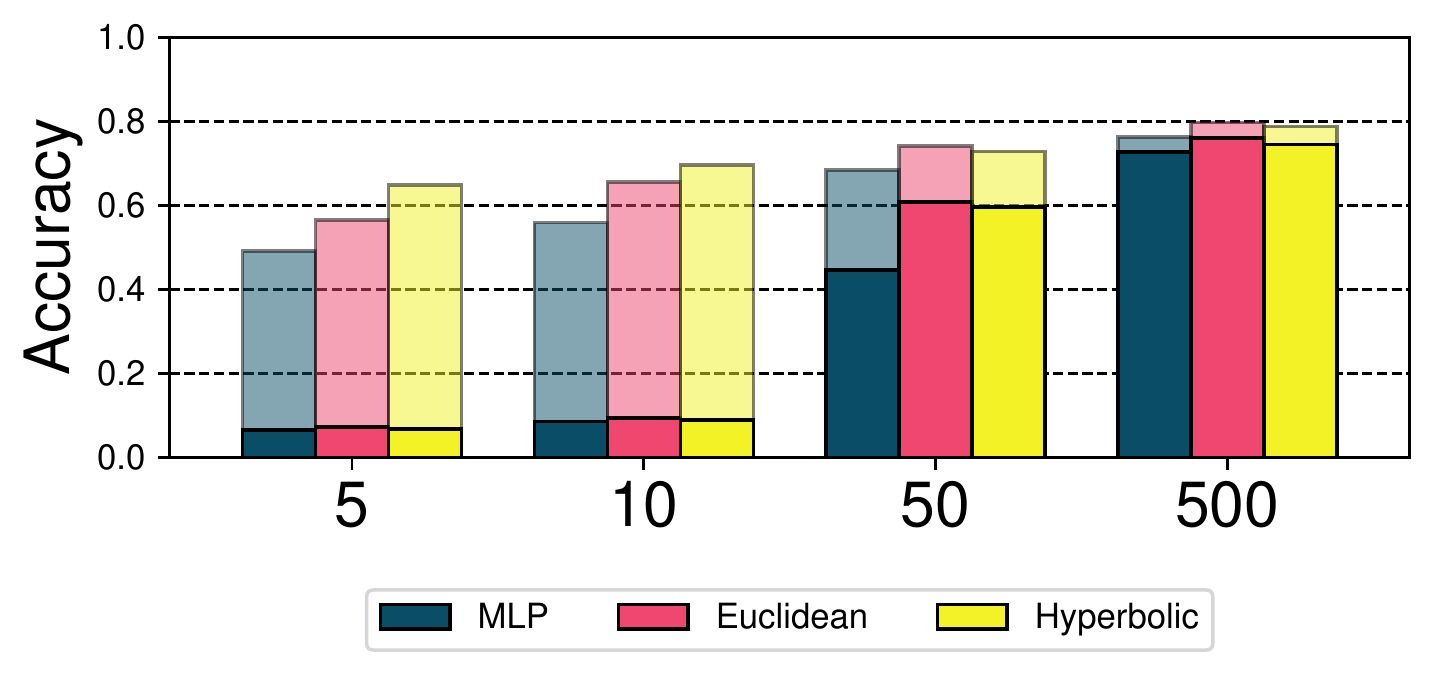}
    \caption{NEWS \label{fig:acc_newsgroup}}
    \end{subfigure}
    \caption{Accuracy gains from pretraining as a function of the number of examples per class available in the new label distribution. While the models are comparable in the pretraining stage (solid bards), the prototypical models make better use of pretraining, showing higher gain during fine-tuning in both the low and high data data regimes (translucent bars).} \label{fig:pretrain_gains_news}
\end{figure}
We wish to isolate the effect of pretraining on an older label set by measuring the performance of our models on the new label distribution with and without pretraining. Figure~\ref{fig:pretrain_gains_news} shows accuracy without pretraining as solid bars, with the gains due to pretraining shown as translucent bars above them. In the low-data regime without pretraining, all models often perform similarly. Nevertheless, our models do improve substantially over the baseline once pretraining is introduced.

With only a few new examples, our models better leverage knowledge gained from old pretraining data. On the NEWS dataset in particular, with only 5 fine-tune examples per class, the relative reduction in classification error for metric learning models exceeds $53\%$ (Euclidean) and $62\%$ (hyperbolic), while the baseline only reduces relative error by about $45\%$. This shows that the prototypical network, and particularly the hyperbolic model can adapt more quickly to dynamic label shifts. Furthermore, the prototypical models conserve their advantage over the baseline in the high data regime, though the margins become smaller.

\begin{table}[!ht]
	\begin{center}
		\begin{tabular}{|l|r|r|r|r|}
			\hline
			Model & 5 &  10 &   20 &  100 \\
			\hline
			MLP (un-tuned) & 38.2 & 46.7 & 42.4 & 46.3  \\
		    EUC & 39.6 & 45.5 & 47.7 & \bf{62.7} \\
		    EUC (un-tuned) & 43.4 & 51.2 & 47.6 & 55.8 \\
			HYP  & 42.2 & 47.1 & 53.0 & \bf{62.7}\\
			HYP (un-tuned)  & \bf{45.7} & \bf{52.4} & \bf{53.3} & 53.1 \\
			 \hline
	    \end{tabular}
		\caption*{(a) SENT}

        \vspace{1em}
        \begin{tabular}{|l|r|r|r|r|}
			\hline
			Model & 5 &  10 & 50 &  500 \\
			\hline
			MLP (un-tuned) & 29.5 & 34.6 & 40.3 & 42.7 \\
		    EUC & 56.5 & 65.6 & \bf{74.2} & \bf{79.8} \\
		    EUC (un-tuned) & 53.2 & 56.5 & 59.6 & 60.8 \\
			HYP & \bf{64.8} & \bf{69.7} & 72.9 & 78.8 \\
			HYP (un-tuned) & 60.1 & 62.9 & 65.4 & 66.7 \\
			 \hline
	    \end{tabular}
	    \caption*{(b) NEWS}
	\end{center}
    \caption{Test accuracy for each dataset and method. Columns indicate the number of examples per label used for fine-tuning and/or creating prototype vectors.}\label{tab:finetune}
\end{table}

\paragraph{Benefits of Fine-tuning}
An important advantage of the prototypical model is its ability to predict classes that were unseen during training with as few as a single support point for the new class. A natural question is whether fine-tuning on these new class labels immediately improves performance, or whether fine-tuning should only be done once a significant amount of data has been obtained from the new distribution. We study this question by comparing the performance of tuned and un-tuned models on the new label distribution.

Table \ref{tab:finetune} compares the accuracy of two types of pretrained prototypical models provided with a variable number of new examples. The fine-tuned model uses this data for both additional training and for constructing new prototypes. The \textit{un-tuned} model constructs prototypes using the pretrained model's representations without additional training. We also construct an un-tuned MLP baseline by fitting a nearest neighbor classifier (KNN, k=5) on the encodings of the penultimate layer of the network. We experimented with fitting the KNN on the output predictions but found that using the penultimate layer was more effective.

We find that the models generally benefit from fine-tuning once a significant amount of data for the new classes is provided ($>$ 20). In the low data regime, however, the results are less consistent, and suggests that the performance may be very dataset dependant. We note however that all metric learning models significantly outperform the MLP-KNN baseline in both the low and high data regimes. This shows that regardless of fine-tuning, our approach is more robust on previously unseen classes.

\begin{figure}[!t!]
    \centering
    \begin{subfigure}{0.9\columnwidth}
    \centering
    \includegraphics[width=\textwidth]{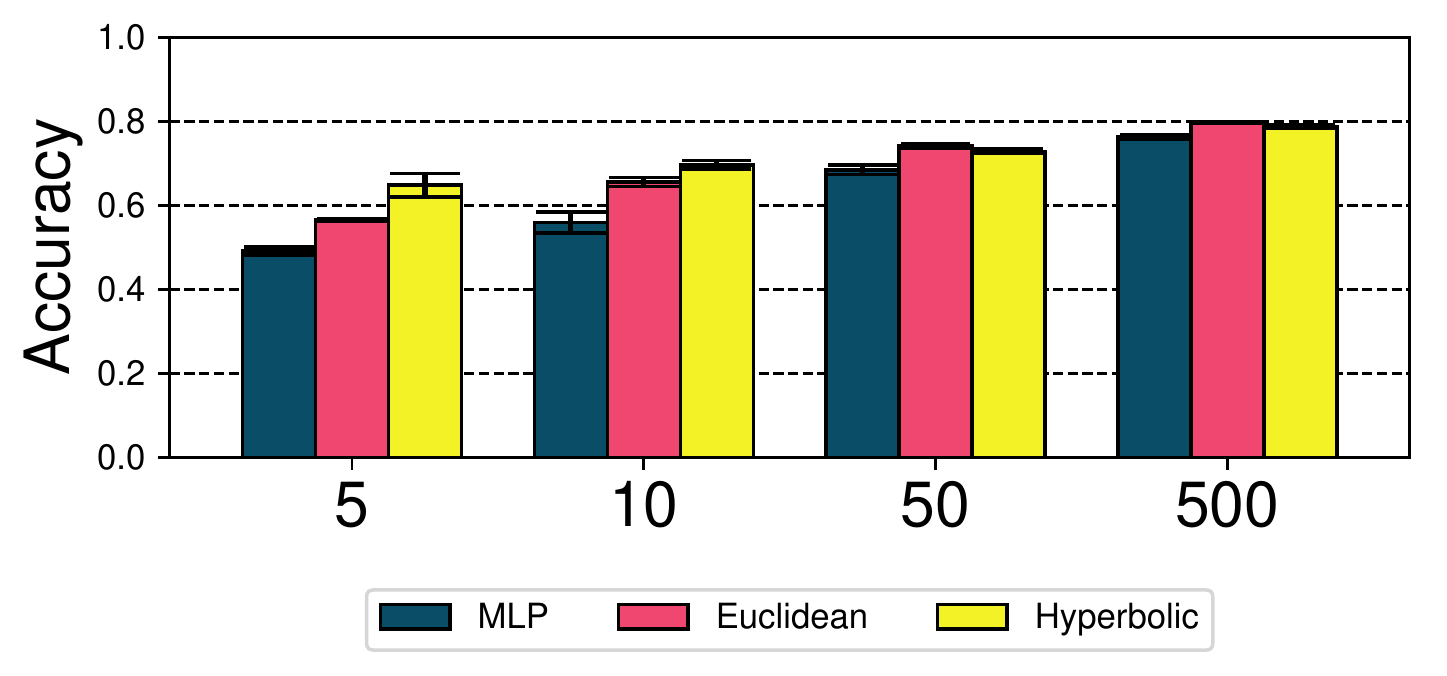}
    \caption{\label{fig:acc_sentiment_all} Accuracy with respect to the full label set}
    \end{subfigure}\\
    \begin{subfigure}{0.9\columnwidth}
    \centering
    \includegraphics[width=\textwidth]{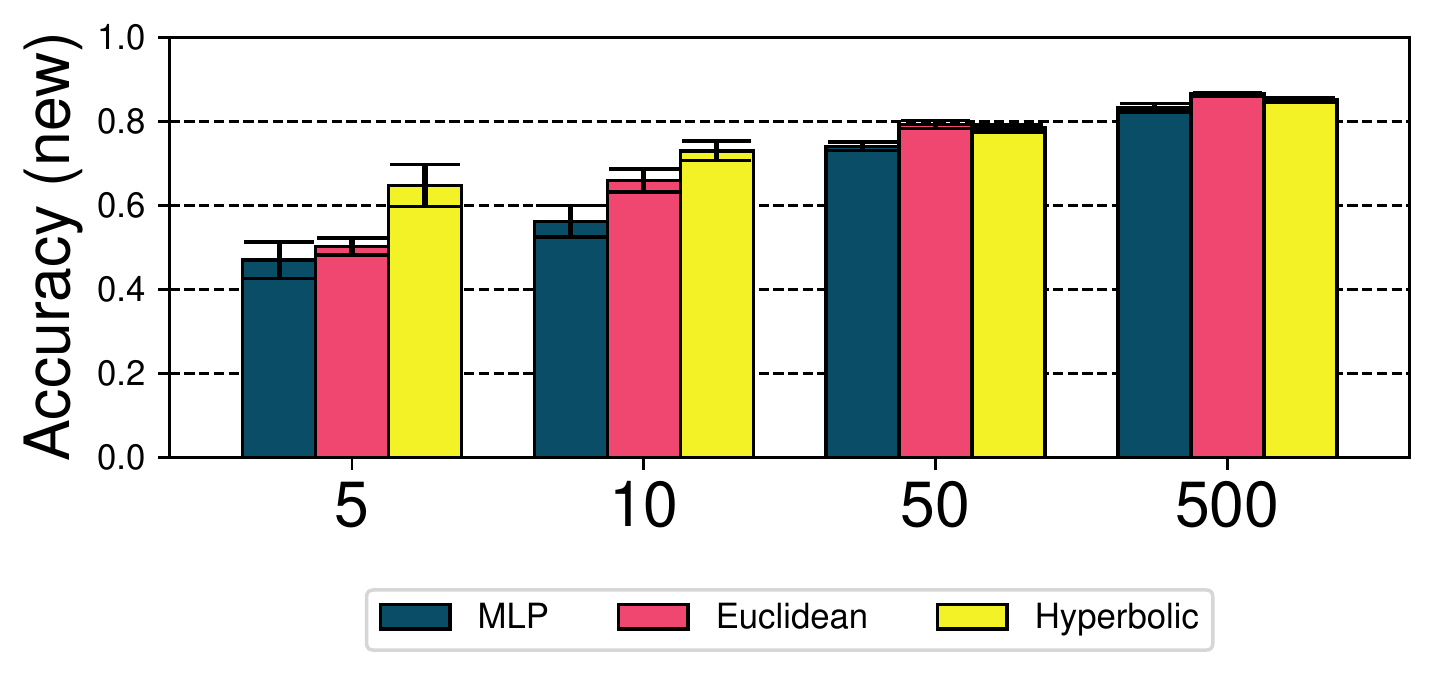}
    \caption{\label{fig:acc_sentiment_new} Accuracy with respect to new classes only}
    \end{subfigure}
    \caption{Accuracy on the NEWS Dataset against number of fine tune examples: (a) all classes and (b) newly introduced classes only. The mean is taken over 5 random label splits, and error bars are given at $\pm 1$ standard deviation.} \label{fig:acc_sentiment}
    \vspace{-0.2in}
\end{figure}

\paragraph{Learning New Classes}\label{ssec:new_classes}
An important factor in the dynamic classification setup is the ability for the model to not only keep performing well on the old classes, but also to smoothly adapt to new ones. We highlight the performance of the models on the newly introduced labels in Figure~\ref{fig:acc_sentiment}, where we see that the improvement in accuracy is dominated by the performance on the new classes.
Plots for additional datasets are shown in Appendix~\ref{sec:app_experiments}.

%% file: conclusions.tex
We propose a framework for dynamic text classification in which the label space is considered flexible and subject to frequent changes. We apply a metric learning method, the prototypical network, and demonstrate its robustness for this task in a variety of data regimes. Motivated by the idea that new labels often originate from label splits, we extend prototypical networks to hyperbolic geometry, derive expressions for hyperbolic prototypes, and demonstrate the effectiveness of our model in the low-resource setting. Our experimental findings suggest that metric learning improves dynamic text classification models, and offer insights on how to combine low-resource training data from overlapping label sets. In the future we hope to explore other applications of metric learning to low-resource research, possibly in combination with explicit models for label entailment (tree learning, fuzzy sets), and/or Wasserstein distance.

%% file: mean_proof.tex
\begin{proposition_nonum}
Every finite collection of points $X$ in $\hypspace^d$ has a unique mean $p^{\star}_{X}$. Furthermore, solving the optimization problem \eqref{eq:mean_function} with Riemannian gradient descent will converge to $p^{\star}_{X}$.
\end{proposition_nonum}

The idea of the proof is to use a known result, which states that our optimization target $\phi_{X}(p)$ is strictly convex under the assumption that $X$ is contained in a compact ball. In order to use this result, we show that any ball in hyperbolic space is geodesically convex.

\begin{proposition}[Proposition 60 in \cite{berger-riemanian}]
Given a manifold $M$, a compact set $A$ contained within a convex ball, and a mass distribution $da$ with total mass one, the function $f: m \in M \rightarrow \frac{1}{2}\int_A d(m, a)^2da$ is strictly convex and achieves a unique minimizer.
\end{proposition}

To use this proposition for our purposes, let the distribution $da$ be the uniform distribution over the points $x \in X$. We must show that $X$ is contained inside a convex ball. Let $c$ be any point in hyperbolic space, and let $R = 2 \max_{x \in X} d_{\hypspace}(c, x)$. Then $X$ is contained within a ball of radius $R$ about $c$, which we will call $V$.

We need to show that $V$ is geodesically convex. In the hyperboloid model, geodesics coincide with the intersection of planes through the origin with the hyperboloid (\citet{cannon-hyperbolic}, page 80).

Given points $u, v \in V$, we need that the geodesic segment between $u$ and $v$ is contained in $V$. This geodesic segment is precisely the projection onto the hyperboloid through the origin of the line segment $t u + (1-t)v$ for $0\le t \le 1$, where the sum is in the ambient real vector space.  This projection is given by $\frac{tu + (1-t)v}{\sqrt{B(t u + (1-t)v, t u + (1-t)v)}}$.

By the hyperbolic distance formula \eqref{eq:hyperboloid_distance}, the ball $V$ of radius $R$ centered on $c$ is precisely all points $x$ in the hyperboloid such that $B(x, c) < \cosh(R)$. Then

\begin{align*}
&B\left(\frac{tu + (1-t)v}{\sqrt{B(tu + (1-t)v,tu + (1-t)v)}}, c\right) \\
&= \frac{B(tu + (1-t)v, c)}{\sqrt{B(tu + (1-t)v,tu + (1-t)v)}}  \\
&\leq B(tu + (1-t)v, c) \\
&= t(B(u, c)) + (1-t)(B(v,c)) \\
&< t\cosh(R) + (1-t) \cosh(R)\\
&= \cosh(R),
\end{align*}

where we have used the fact that the normalizing constant $\sqrt{B(tu + (1-t)v,tu + (1-t)v)}$ is greater than or equal to $1$, the linearity of $B(\cdot, \cdot)$, and the fact that endpoints $u$ and $v$ are in $V$.
Hence the geodesic segment is completely contained in $V$, and the proposition gives us strict convexity of $\phi$, as well as existence and uniqueness of $p^{\star}$. Strict convexity of a smooth function on a compact ball gives strong convexity on a slightly smaller ball, smoothness on a compact ball gives Lipschitz continuity of the gradient, and hyperbolic space has constant curvature $-1$, making it a Hadamard manifold, hence Riemannian gradient descent will converge to a minimum assuming a sufficiently small step size (\cite{zhang-sra-geodgraddesc}, theorem 13).

\qedsymbol

%% file: means_differ_example.tex
Here we show that the mean $p^{\star}_{X}$ given by \eqref{eq:mean_function} differs from that of \eqref{eq:closed_form_mean}. Let $a = [1, 0, 0]$ and $b = [\sqrt{2}, 1, 0]$.  Let $X = \{a, a, b\}$.  Then both these means lie on the geodesic between $a$ and $b$.  We can compute the hyperbolic distance between $a$ and $b$ via \eqref{eq:hyperboloid_distance}:
\begin{align*}
d = \arccosh(B(a,b)) = \arccosh(\sqrt{2})
\end{align*}
Then the minimizer of $\phi_X$ is the point on the geodesic at distance $\frac{\arccosh(\sqrt{2})}{3} = 0.2938$ from $a$.  Utilizing \eqref{eq:closed_form_mean} we get:
\begin{align*}
\tilde{p} = \frac{1}{\sqrt{5 + 4\sqrt{2}}} [2 + \sqrt{2}, 1, 0] .
\end{align*}
Computing the distance from $a$ to $\tilde{p}$ using \eqref{eq:hyperboloid_distance} we get:
\begin{align*}
d_{\hypspace}(a, \tilde{p})  = \arccosh\left(\frac{2 + \sqrt{2}}{\sqrt{5 + 4\sqrt{2}}}\right) = 0.3017 .
\end{align*}

Hence the two means are not equivalent.

Note that the data set we used to provide this example involves a very skewed set, in which twice as many points are on one side as the other, yet the difference between the two is less than $3\%$. We conjecture that the difference can be bounded tightly by a function of $|X|$ and the diameter of its convex hull; proving this claim is an interesting open question for future work.

%% file: hierarchy_appendix.tex
\begin{verbatim}
ROOT:  
  COMPUTER:
    GRAPHICS (leaf)
    WINDOWS:
      WINDOWSX: (leaf)
      WINDOWSOS (leaf)
    HARDWARE:
      PC (leaf)
      MAC (leaf)
  RECREATIONAL:
    VEHICLES:
      AUTOS (leaf)
      MOTORCYCLES (leaf)
    SPORTS:
      BASEBALL (leaf)
      HOCKEY (leaf)
  SCIENCE:
    CRYPT (leaf)
    ELECTRONICS (leaf)
    MEDECIN (leaf)
    SPACE (leaf)
  POLITICS:
    GUNS (leaf)
    MIDEAST (leaf)
    POLITICSMISC (leaf)
  RELIGION:
    CHRISTIAN (leaf)
    ATHEISM (leaf)
    RELIGIONMISC (leaf)
  FORSALE (leaf)
\end{verbatim}

%% file: appendix_datasets.tex
Here we present additional details for the datasets used in our experiments.

\begin{table*}[t!]
\begin{center}
\begin{tabular}{|l|r|r|r|r|r|}
\hline \bf Name & \bf \# Train & \bf \# Dev & \bf \# Test & \bf \# Labels & \bf Depth   \\ \hline
$20$ Newsgroups (NEWS) & $9,\!044$ & $2,\!668$ & $7,\!531$ & $20$ & $3$  \\
Web of Science (WOS) & $25,\!182$ & $6,\!295$ & $15,\!503$ & $134$ & $2$  \\
Twitter Airline Sentiment (SENT) & $5,\!975$ & $742$ & $754$ & $10$ & Flat  \\
\hline
\end{tabular}
\end{center}
\caption{\label{tab:data} Datasets for Sections~\ref{sec:experiments}--\ref{sec:analysis}. Half the training documents are used for pretraining on $\mathcal{Y}_{old}$.}
\end{table*}

%% file: extra_experiments.tex
Here we present results from the analysis experiments of Sections~\ref{sec:experiments} and~\ref{sec:analysis} on additional datasets.

\begin{figure*}[!t!]
    \centering
    \begin{subfigure}{0.9\columnwidth}
    \centering
    \includegraphics[width=\textwidth]{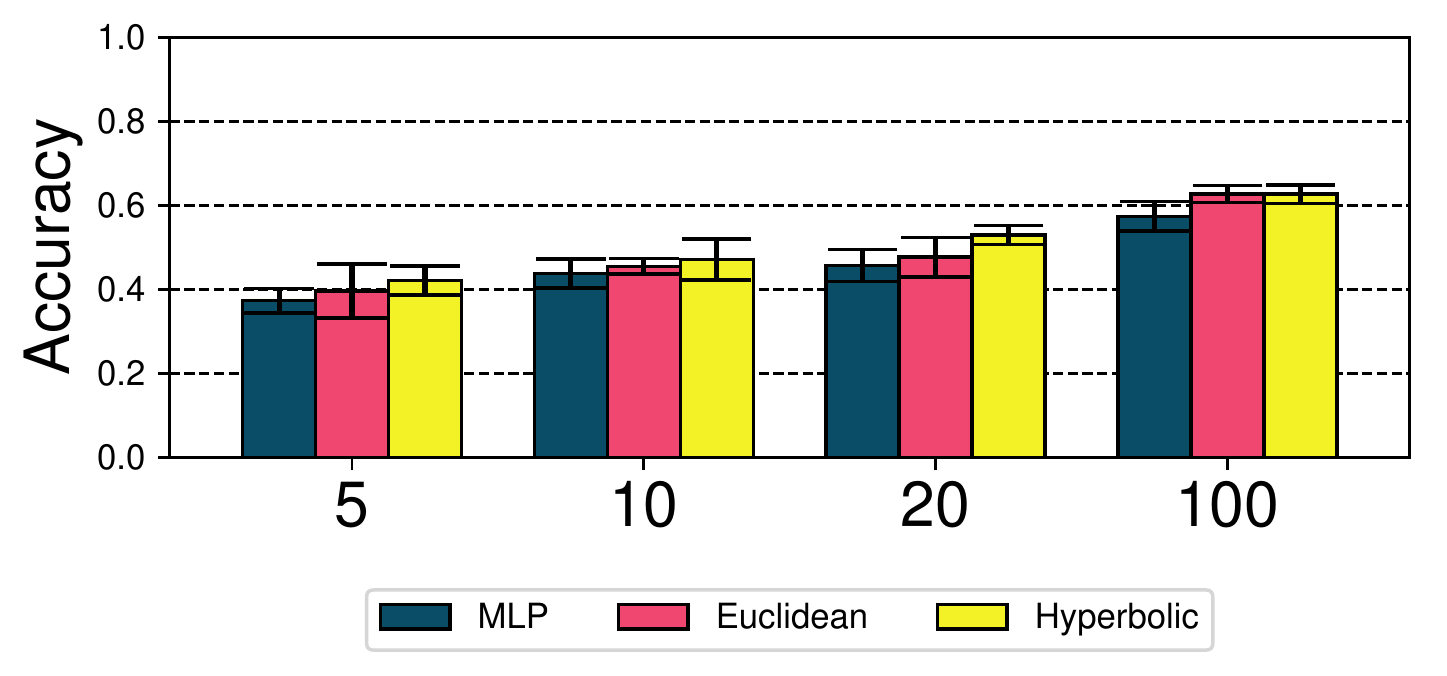}
    \caption{\label{fig:acc_extra2_all}}
    \end{subfigure}
    \begin{subfigure}{0.9\columnwidth}
    \centering
    \includegraphics[width=\textwidth]{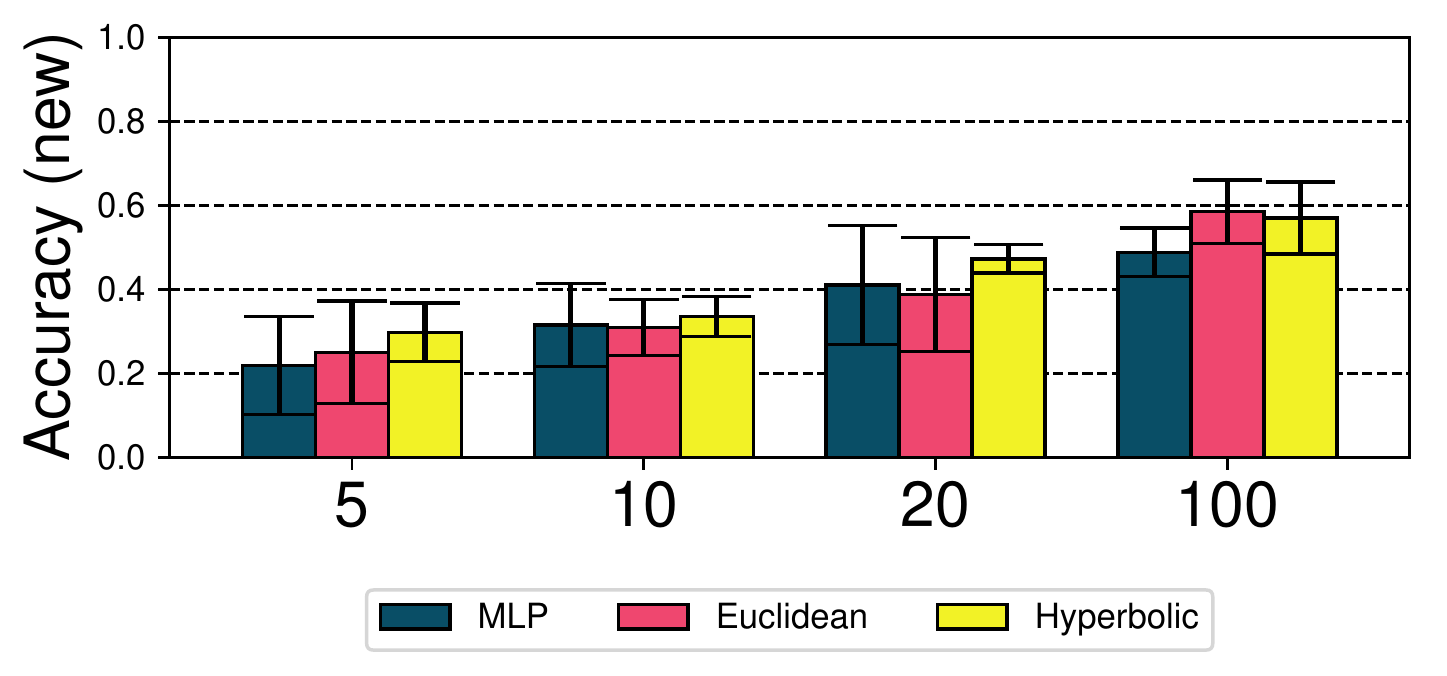}
    \caption{\label{fig:acc_extra2_new}}
    \end{subfigure}
    \caption{Accuracy on the SENT Dataset against number of fine tune examples: (a) all classes and (b) newly introduced classes only. The mean is taken over 5 random label splits, and error bars are given at $\pm 1$ standard deviation.} \label{fig:acc_extra2}
\end{figure*}

\begin{figure*}[!t!]
    \centering
    \begin{subfigure}{0.9\columnwidth}
    \centering
    \includegraphics[width=\textwidth]{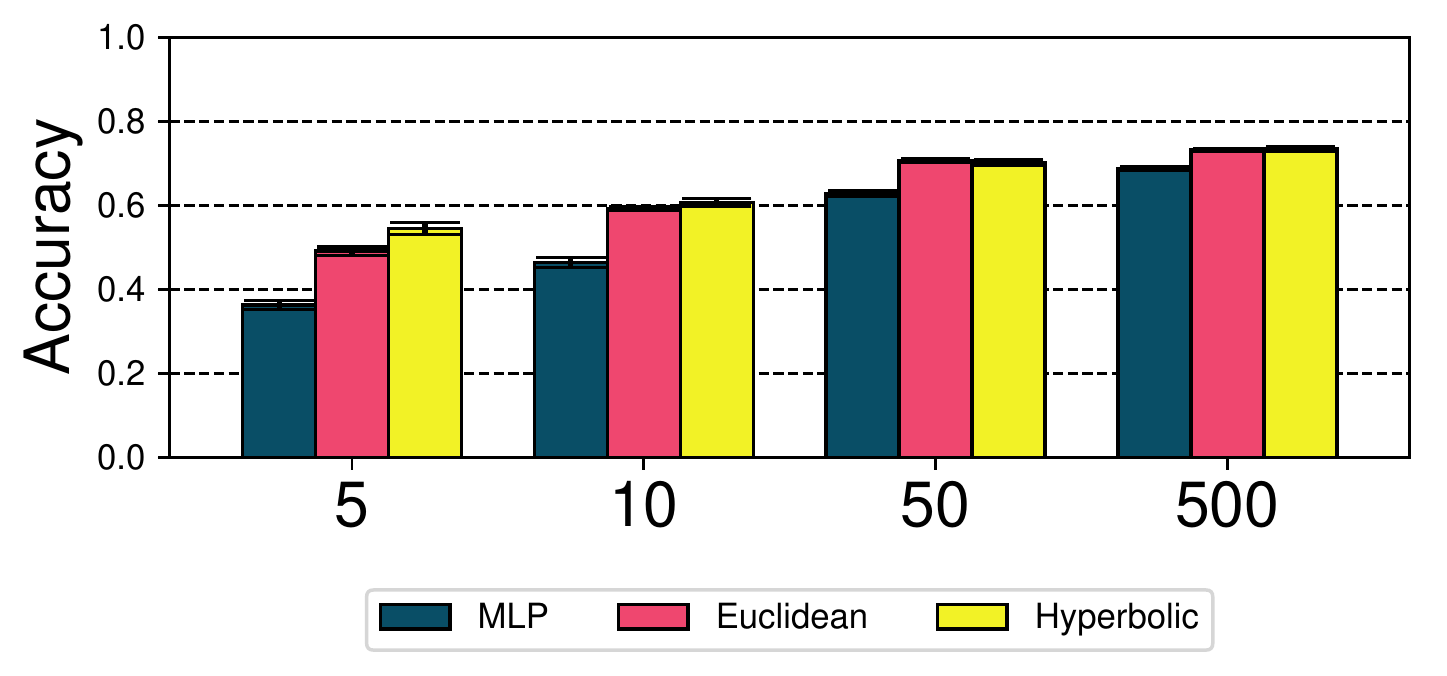}
    \caption{\label{fig:acc_extra3_all}}
    \end{subfigure}
    \begin{subfigure}{0.9\columnwidth}
    \centering
    \includegraphics[width=\textwidth]{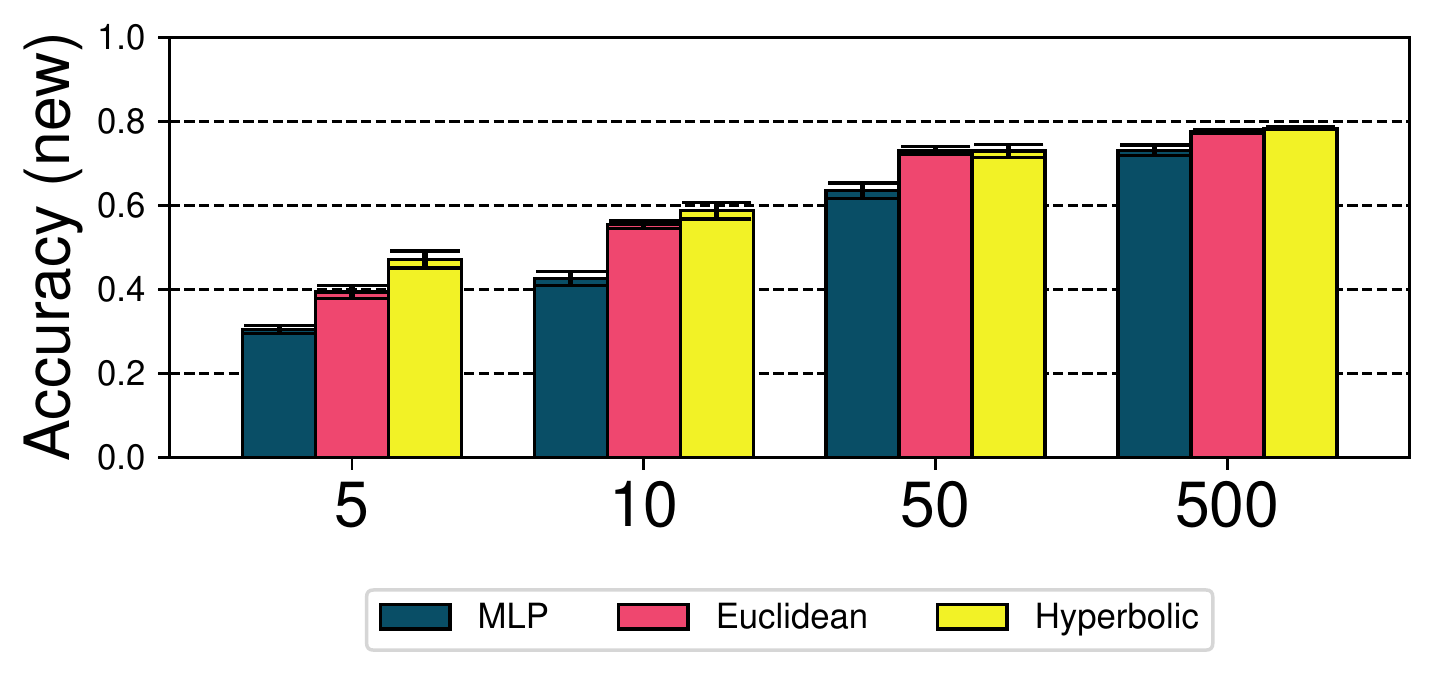}
    \caption{\label{fig:acc_extra3_new}}
    \end{subfigure}
    \caption{Accuracy on the WOS Dataset against number of fine tune examples: (a) all classes and (b) newly introduced classes only. The mean is taken over 5 random label splits, and error bars are given at $\pm 1$ standard deviation.} \label{fig:acc_extra3}
\end{figure*}